\documentclass[letterpaper]{article} % DO NOT CHANGE THIS
\usepackage{aaai2026}
\usepackage{times}  % DO NOT CHANGE THIS
\usepackage{helvet}  % DO NOT CHANGE THIS
\usepackage{courier}  % DO NOT CHANGE THIS
\usepackage[hyphens]{url}  % DO NOT CHANGE THIS
\usepackage{graphicx} % DO NOT CHANGE THIS
\usepackage{natbib}  % DO NOT CHANGE THIS AND DO NOT ADD ANY OPTIONS TO IT
\usepackage{caption} % DO NOT CHANGE THIS AND DO NOT ADD ANY OPTIONS TO IT
\usepackage{algorithm}
\usepackage{algorithmic}

\usepackage{newfloat}
\usepackage{listings}
\DeclareCaptionStyle{ruled}{labelfont=normalfont,labelsep=colon,strut=off} % DO NOT CHANGE THIS
\floatstyle{ruled}
\newfloat{listing}{tb}{lst}{}
\floatname{listing}{Listing}
\usepackage{amsmath,amssymb}
\usepackage{bm}
\usepackage{amsfonts}
\usepackage{graphicx}
\usepackage{booktabs}
\usepackage{multirow}
\usepackage{multicol}
\usepackage{adjustbox}
\usepackage{colortbl}

\title{RoadMamba: A Dual‑Branch Visual State Space Model \\ for Road Surface Classification}
\author{
    Tianze Wang \equalcontrib, 
    Zhang Zhang \equalcontrib, 
    Chao Yue, Nuoran Li, Chao Sun $^{\dagger}$
}

\affiliations{
    \textsuperscript{\rm }Beijing Institute of Technology\\
    chaosun@bit.edu.cn
}

\usepackage{bibentry}
\begin{document}

\maketitle

\begin{abstract}
Acquiring the road surface conditions in advance based on visual technologies provides effective information for the planning and control system of autonomous vehicles, thus improving the safety and driving comfort of the vehicles. Recently, the Mamba architecture based on state-space models has shown remarkable performance in visual processing tasks, benefiting from the efficient global receptive field. However, existing Mamba architectures struggle to achieve state-of-the-art visual road surface classification due to their lack of effective extraction of the local texture of the road surface. In this paper, we explore for the first time the potential of visual Mamba architectures for road surface classification task and propose a method that effectively combines local and global perception, called RoadMamba. Specifically, we utilize the Dual State Space Model (DualSSM) to effectively extract the global semantics and local texture of the road surface and decode and fuse the dual features through the Dual Attention Fusion (DAF). In addition, we propose a dual auxiliary loss to explicitly constrain dual branches, preventing the network from relying only on global semantic information from the deep large receptive field and ignoring the local texture. The proposed RoadMamba achieves the state-of-the-art performance in experiments on a large-scale road surface classification dataset containing 1 million samples.
\end{abstract}


\section{Introduction}
Autonomous vehicles \cite{geiger2013vision}, \cite{caesar2020nuscenes}, \cite{sun2020scalability} and intelligent transportation system \cite{yu2022dair, zhang2025height3d, yang2023bevheight, zhang2025heightformer} have been widely explored and developed in recent years, making vehicle safety and ride comfort become increasingly important. Accurate road surface information, including friction levels, uneven conditions and more, helps driving system control their vehicles within a safe and comfortable range. As cameras are becoming the standard for driving environment sensing, many studies \cite{nolte2018assessment, leng2021estimation, tian2021reliable, dhiman2019pothole, vsabanovivc2020identification, zhao2023comprehensive, wang2025roadformer} have verified that vision-based road surface perception is one of the effective and superior methods. Specifically, it performs fine-grained classification of road surface images collected by the on-board camera using Deep Neural Network (DNN), such as the friction level, unevenness, and material properties of the road surface. Among them, the subclasses of the three road properties are combined to model it as a classification task with 27 classes. Fine-grained road surface classification with different combinations of road attributes requires the network not only to be able to obtain accurate global semantics from the road, but also not to ignore the local nuances of the surface texture, which pose a challenge to existing image classification networks. For example, the material and friction level of the road surface are mainly obtained from the global semantics of the image, while the damage and unevenness level of the road surface are mainly recognized from some local details.

Recently, visual Mamba \cite{gu2023mamba, zhu2024vision, liu2024vmamba, guo2024mambair, chen2025q, zhang2025pillarmamba} based on State Space Model (SSM) has emerged as an efficient and effective backbone for constructing deep networks. This development offers a potential solution for efficient long-range modeling in road surface classification. Specifically, the discretized recursive state-space equations and the parallel scanning algorithm in Mamba can efficiently model very long-range dependencies, striking a balance between global receptive fields and computational burden compared to the standard Transformer. This implies that Mamba-based classification methods have a natural advantage in efficiently processing global pixels of road surface images and obtaining global semantics. The above desirable properties motivate us to explore the potential of visual Mamba for road surface classification.

However, existing visual Mamba architectures still face challenges when dealing with the combination of road surface properties mentioned above for road surface classification due to the lack of effective extraction of local texture. Since standard Mamba was designed to handle 1-D sequences in Natural Language Processing (NLP), it naturally faces the separation of neighboring pixels when processing 2-D signals. Although recent methods \cite{huang2024localmamba, liu2024vmamba} have attempted to improve on the scanning direction and to some extent increase the modeling of local 2D dependencies, however, the local details still run the risk of being forgetting in long sequences.

To address this, we propose a method that effectively combines local and global perception, called RoadMamba. Specifically, we propose the Dual State Space Model (DualSSM) consisting of Local State Space Model (LocalSSM) branch and Global State Space Model (GlobalSSM) branch to efficiently extract the global semantics and local texture of road surfaces. To further decode and fuse the extracted global and local features, Dual Attention Fusion (DAF) is proposed. It utilizes channel attention to extract key semantic information and suppress irrelevant channels in global semantics, and retains key spatial distributions in local information through spatial attention. In addition, we propose an auxiliary loss to explicitly constrain dual branches, preventing the network from relying only on global semantic information from the deep large receptive field and ignoring the local texture.

In short, our main contributions can be summarized as follows:
 \begin{itemize}
\item We explore for the first time the potential of visual Mamba architectures for road surface classification task and propose a framework that effectively extracts global semantics and local textures from road surface based on Dual State Space Model (DualSSM), called RoadMamba.\\

\item In order to effectively decode and fuse the extracted global and local features, we propose the Dual Attention Fusion (DAF), which consists of lightweight global and local attention to extract key semantic and retain critical spatial distributions.\\

\item We utilize a dual auxiliary loss to explicitly constrain dual branches, preventing the network from relying only on global semantic information from the deep large receptive field and ignoring the local texture.\\

\item Extensive experiments on a large-scale road surface classification dataset containing 1 million samples demonstrate our RoadMamba outperforms the state-of-the-art methods.

\end{itemize}

\section{Related Work}
\subsection{Road Surface Classification}
The main technical route for vision-based road perception is the utilization of Deep Neural Networks (DNNs) to classify road categories or to detect road unevenness. \cite{roychowdhury2018machine} developed a classification scheme that takes into account the characteristics of the drivable surface, the sky, and the surrounding environment. \cite{vsabanovivc2020identification} designed a DNN-based algorithm to classify three road surfaces under dry and wet conditions. In recent years, \cite{zhao2023comprehensive} has collected and created datasets of road surface images from real-world driving conditions that scale up to 1 million samples, labeling road images in detail with friction levels, unevenness, and material properties, which made significant advancements in the road surface classification. On this basis, RoadFormer \cite{wang2025roadformer} achieved effective road surface classification by combining the architectures of CNN and Transformer. However, it relies on specific combination strategies and architectural designs that do not realize the potential of global-local feature extraction.

\subsection{State Space Model}
State Space Model \cite{ssm}, derived from classical control theory, has recently been introduced to deep learning \cite{s4}, \cite{s5}, gaining increasing attention. Since the introduction of Mamba \cite{gu2023mamba}, a number of efforts have been proposed to leverage its capability for vision applications. Vim \cite{zhu2024vision} introduces a bi-directional SSM scheme that processes tokens in the forward and backward directions to capture global context and improve spatial comprehension. VMamba \cite{liu2024vmamba} proposes a Cross-Scan module. This module uses a four-way selective scan methodology to integrate information from surrounding tokens and capture the global context. LocalMamba \cite{huang2024localmamba} has attempted to improve on the scanning direction and to some extent increase the modeling of local 2D dependencies. However, existing visual Mamba architectures still face challenges when dealing with the combination of road surface properties mentioned above for road surface classification due to the lack of effective extraction of local texture. 

\section{Methodology}

\subsection{Preliminaries}

\subsubsection{Structured State Space Models}

Structured State Space Models (SSMs) provide a principled framework for modeling long-range dependencies in sequential data by introducing a latent state that evolves according to linear dynamics. In continuous time, an SSM is defined by
\begin{equation}\label{eq:ssm_continuous}
\begin{aligned}
h'(t) = \boldsymbol{A} h(t) + \boldsymbol{B} x(t), 
y(t) = \boldsymbol{C} h(t)
\end{aligned}
\end{equation}
where \(x(t)\in\mathbb{R}^L\) is the input sequence, \(h(t)\in\mathbb{R}^N\) is the hidden state, \(y(t)\in\mathbb{R}^L\) is the output, and
\(A\in\mathbb{R}^{N\times N}\), \(B\in\mathbb{R}^{N\times L}\), \(C\in\mathbb{R}^{L\times N}\)
govern the system dynamics and readout.

To implement SSMs in discrete-time deep learning frameworks, one commonly applies a zero-order hold (ZOH) discretization over a time step \(\Delta\), yielding
\begin{equation}\label{eq:zoh}
\begin{aligned}
	\boldsymbol{\bar{A}} = e^{\boldsymbol{\Delta A}},
	\boldsymbol{\bar{B}} = (\boldsymbol{\Delta A})^{-1} (e^{\boldsymbol{\Delta A}} - I) \cdot \boldsymbol{\Delta B}
\end{aligned}
\end{equation}
so that
\begin{equation}\label{eq:ssm_discrete}
	\begin{aligned}
		h_t = \boldsymbol{\overline{A}}{h}_{t-1} + \boldsymbol{\overline{B}}{x}_t,
		{y_t} = \boldsymbol{C}{h}_t
	\end{aligned}
\end{equation} 
Equation \ref{eq:ssm_discrete} can be executed efficiently via associative scans and further recast as a global convolution
\begin{equation}\label{eq:global_conv}
	\begin{aligned}
		\boldsymbol{y} = \boldsymbol{x} \;\circledast\;\overline{\boldsymbol{K}}~~~~~~~~~~~~~~~~~~~~~~\\
		\overline{\bm{K}}
		= \bigl(
		\bm{C}\,\overline{\bm{B}},\;
		\bm{C}\,\overline{\bm{AB}},\;
		\dots,\;
		\bm{C}\,\overline{\bm{A}}^{L-1}\,\overline{\bm{B}}
		\bigr)
	\end{aligned}
\end{equation}
where $\circledast$ denotes convolution and $\overline{\boldsymbol{K}} \in \mathbb{R}^{N\times L}$ is the structured kernel.

\subsubsection{Selective State Space Models}

While classical SSMs (often called S4) scale linearly with sequence length, their use of fixed parameters limits adaptability to varying inputs. Selective State Space Models—exemplified by Mamba—enhance this framework by making the transition and output matrices input-dependent. Specifically, at each time step \(t\), the parameters $\boldsymbol{B_t}$, $\boldsymbol{C_t}$, and the discretization $\boldsymbol{\Delta_t}$ are generated dynamically from the input \(x_t\), enabling a sequence-aware parameterization that filters out irrelevant information and focuses computation on salient features.

\begin{figure*}[th]
    \centering
    \includegraphics[width=\textwidth]{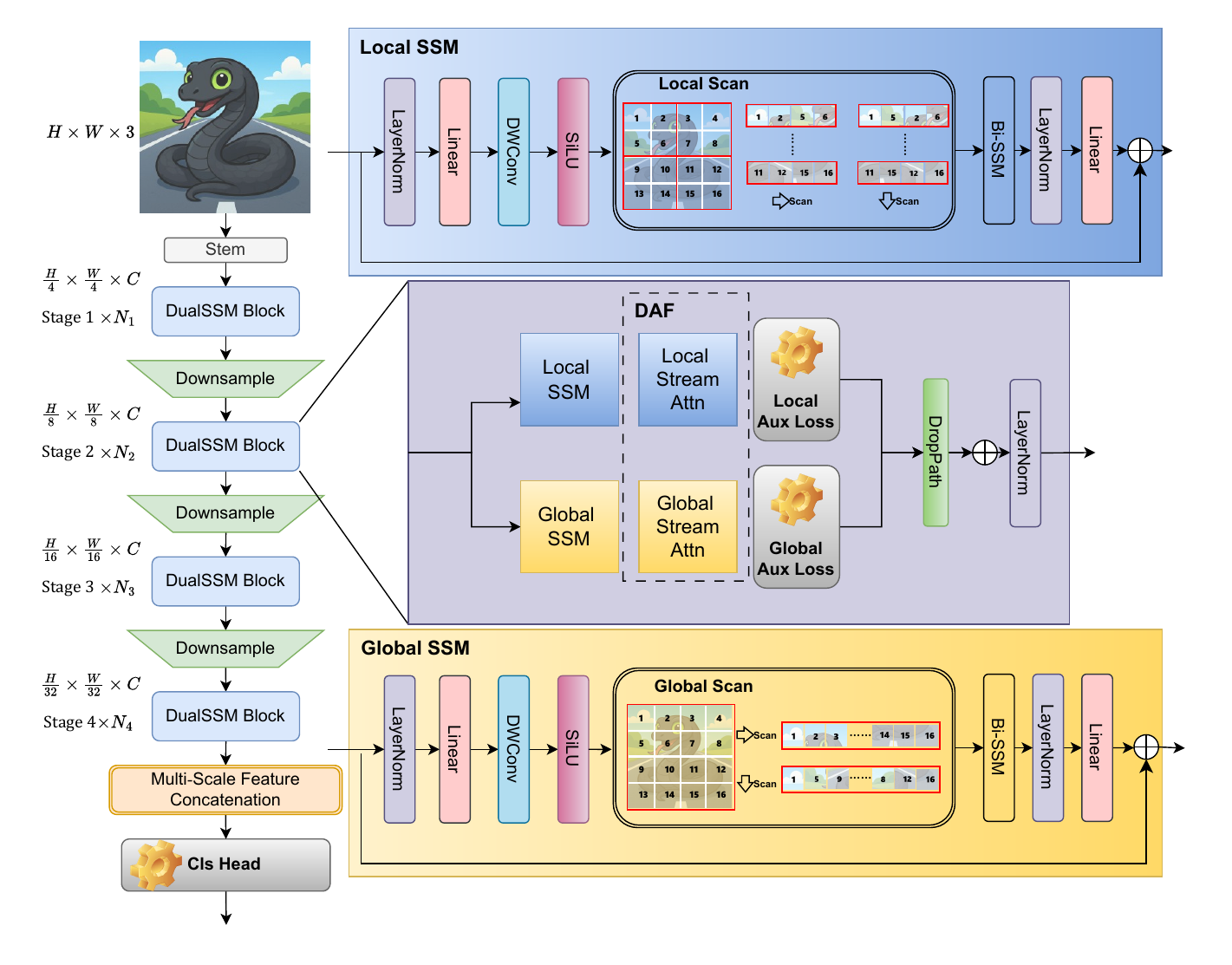}
    \caption{\textbf{Architectural details of the RoadMamba.} The model processes the input image through a stem and four stages of DualSSM Blocks, each containing a Local/Global SSM for parallel extraction of global and local features and a Dual Attention Fusion module to enhance extracted local and global features with auxiliary losses. Multi-scale features from all stages are concatenated and fed into the classification head for final prediction. }
    \label{overall}
\end{figure*}

% \begin{figure*}[th]
% 	\centering
% 	\begin{equation}
% 		\mathbf{x}_h = \text{vec}_{\text{row}}(\mathbf{X}) = [X(1,1,:), X(1,2,:), \ldots, X(1,W,:), X(2,1,:), \ldots, X(H,W,:)] \in \mathbb{R}^{(HW) \times D} \label{xh}
% 	\end{equation}
% \end{figure*}

% \begin{figure*}[th]
% 	\centering
% 	\begin{equation}
% 		\mathbf{x}_{v} = \operatorname{vec}_{\text{col}}(\mathbf{X}) = [X(1,1,:), X(2,1,:), \ldots, X(H,1,:), X(1,2,:), \ldots, X(H,W,:)] \in \mathbb{R}^{(HW) \times D} \label{xv}
% 	\end{equation}
% \end{figure*}

\subsection{Overall Architecture}

Figure~\ref{overall} illustrates the overall architecture of the RoadMamba model. The model adopts a typical hierarchical design, consisting of a Stem module at the input end for extracting initial low-level features, followed by multiple subsequent stages (Stage 1 to Stage 4) that progressively extract higher-level representations. Within Stages 1 to 4, each stage comprises several DualSSM modules, and the resolution of the feature maps is reduced at the end of each stage via 2× downsampling (Patch Merging) while simultaneously increasing the channel dimension. This process results in a pyramid-like multi-scale feature representation. Such an architecture, similar to most visual models for image classification, captures semantic information at different scales by gradually decreasing spatial resolution through successive layers, while effectively controlling computational costs. 

After Stage 4, the network performs multi-scale fusion of the features output by each stage. Specifically, at the end of each stage, we retain the output feature map $F_i\in\mathbb{R}^{H/2^i\times W/2^i\times C_i}$ ($i=1,2,3,4$), which is then normalized using LayerNorm and aggregated into a channel vector via adaptive average pooling:

\begin{equation}
	v_{i} = \operatorname{GAP}(\operatorname{LayerNorm}(F_{i})) \in \mathbb{R}^{C_{i}} \label{vi}
\end{equation}

where $\mathrm{GAP}$ denotes global average pooling, resulting in a vector of shape $1\times1\times C_i$. The channel vectors from the four stages, ${v_1, v_2, v_3, v_4}$, are concatenated along the channel dimension to obtain the fused multi-scale feature representation:

\begin{equation}
	v_{\mathrm{ms}}=\left[v_{1} \| v_{2} \| v_{3} \| v_{4}\right] \in \mathbb{R}^{\sum_{i=1}^{4} C_{i}} \label{vms}
\end{equation}

where “$\|$” denotes the concatenation operation. This multi-scale feature fusion strategy effectively integrates fine-grained texture, edge information, and deep semantic features extracted at different resolutions, providing richer contextual information for subsequent classification. Finally, the concatenated multi-scale feature vector $v_{\mathrm{ms}}$ is fed into the classification head, where a linear layer maps the total channel dimension to the number of classes $N_{\mathrm{cls}}$:

\begin{equation}
	\hat{y} = \text{Linear}\bigl(v_{\text{ms}}\bigr) \in \mathbb{R}^{N_{\text{cls}}} \label{yhat}
\end{equation}

It is worth noting that the entire multi-scale concatenation and classification head are executed only once at the “end” of the model. This design does not alter the computation flow within each stage or increase the intermediate computational load, thus having a negligible impact on the model’s inference time. Nevertheless, it can significantly improve the final classification performance.

At the heart of RoadMamba lies its customized DualSSM module (which will be detailed in the next subsection), a fundamental unit that integrates both global and local scanning capabilities. Within each DualSSM module, the model extracts global and local features in parallel, and then fuses them through a dual-attention fusion mechanism, enabling the output of each stage to simultaneously capture global contextual information and local details. Residual connections are applied between DualSSM modules. With this design, RoadMamba effectively integrates information while maintaining a deep network and strong representational capacity, thereby providing richer feature representations for downstream tasks.

\begin{figure}[t]
    \centering
    \includegraphics[width=0.5\textwidth]{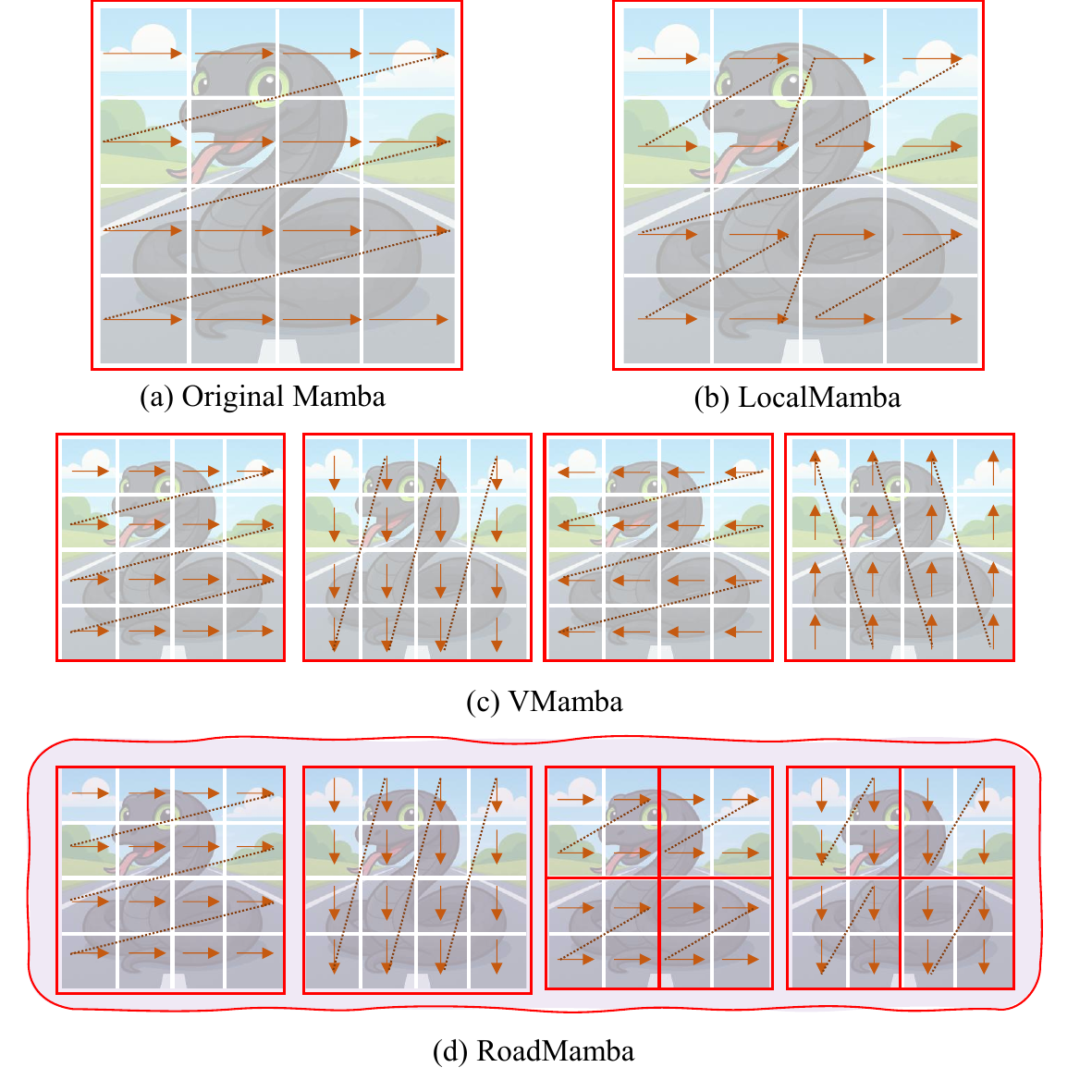}
    \caption{\textbf{Illustration of different scanning strategies.} (a) Original Mamba performs global sequential scanning over the entire image. (b) LocalMamba conducts scanning within local windows and serially connects different windows. (c) VMamba extends scanning directions by applying horizontal and vertical scans within windows. (d) The proposed RoadMamba performs bidirectional scanning separately in global and local windows, where each window operates independently. This explicitly differentiates the roles of global and local feature extraction, further enhancing feature representation capability. }
    \label{scan}
\end{figure}

\subsection{DualSSM Block}

The DualSSM module serves as the fundamental building block of RoadMamba, with its main innovation lying in the integration of both global and local selective state space model (Selective SSM) scanning mechanisms for feature extraction. As illustrated in Figure~\ref{scan}, each DualSSM module contains parallel branches for global scanning and local scanning.

The global scanning branch is responsible for capturing long-range dependencies across the entire feature map. Given an input feature map $X \in \mathbb{R}^{H\times W \times D}$ (where $H$ and $W$ denote the height and width, and $D$ is the number of channels), we flatten the entire feature map along two orthogonal directions into two ultra-long one-dimensional sequences for state space modeling. Specifically, for an input feature tensor $\mathbf{X}$ of shape $H\times W\times D$, we first apply a row-major flattening operation, then a column-major flattening operation is applied to obtain the second sequence.

These two sequences are then fed into one-dimensional SSM units with shared weights. After constant-time stepping and convolutional state updates, we obtain output sequences $\mathbf{y}_h,\mathbf{y}_v\in\mathbb{R}^{(HW)\times D}$. The processed vectors are subsequently reconstructed back into two-dimensional feature maps: specifically, $\mathbf{y}_h$ and $\mathbf{y}_v$ are reshaped into tensors $Y_h$ and $Y_v$ of size $H\times W\times D$, respectively. Finally, these tensors are summed element-wise at each position to produce the global scanning output:

\begin{equation}
	Y_{\text{global}} = Y_h + Y_v \label{yg}
\end{equation}

With this global scanning approach, RoadMamba is able to perceive and model long-range dependencies between any two points in the image within a single forward pass, without being limited by local receptive fields. At the same time, owing to the linear time complexity of SSM, this operation significantly enhances the model’s capacity to capture global context while maintaining high computational efficiency.

The local scanning branch is designed to capture fine-grained local textures and high-frequency details. The input feature map is divided into a grid of non-overlapping windows of size $M\times M$ (with $M$ set to 7 by default), resulting in $n_H \times n_W$ windows, where $n_H=\lceil H/M\rceil$ and $n_W=\lceil W/M\rceil$. For each window block $X_{p,q} \in \mathbb{R}^{M\times M \times D}$ ($0\le p < n_H, 0\le q < n_W$), we perform both horizontal and vertical SSM scans to obtain the local output $Y_{p,q}^{local}$. However, unlike the global branch, to reduce computational cost, we do not process all windows. Instead, only half of the windows are randomly selected for SSM scanning, while the outputs of the remaining windows are set to zero. Let $\mathcal{S}$ denote the index set of the selected windows, then:

\begin{equation}
	Y_{p,q}^{local} = \begin{cases}
		SSM_h(\cdot) + SSM_v(\cdot) & (p,q) \in \mathcal{S} \\
		0 & (p,q) \notin \mathcal{S}
	\end{cases} \label{ypq}
\end{equation}

where $SSM_h$ and $SSM_v$ denote the horizontal and vertical SSM transformations applied to each window, respectively. Finally, all local outputs are reassembled to form a feature map $Y_{\text{local}}\in \mathbb{R}^{H\times W \times D}$ with the same spatial dimensions as the input. Since only about half of the windows undergo local scanning, this strategy significantly reduces computational cost and introduces a regularization effect during training similar to Dropout. The incorporation of the local branch enables the model to focus on fine-grained patterns within local regions, thereby complementing the global branch's limited capacity for capturing detailed information.

In the DualSSM module, the global scanning features $Y_{\text{global}}$ and the local scanning features $Y_{\text{local}}$ are integrated by a subsequent Dual Attention Fusion module (which will be detailed in the next subsection), thereby fully leveraging their complementary information. With this design, each DualSSM module is capable of modeling both global semantics and local textures simultaneously, greatly enhancing the richness and diversity of feature representations.

\begin{figure}[t]
    \centering
    \includegraphics[width=0.5\textwidth]{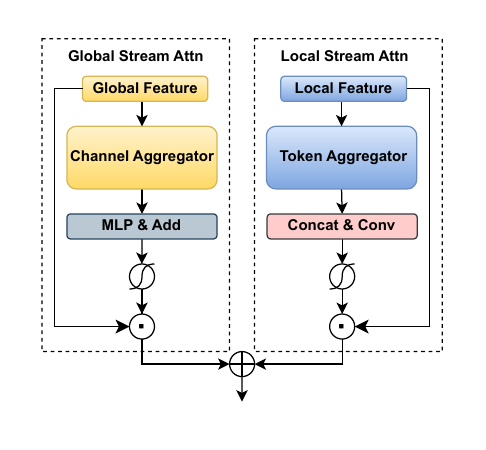}
    \caption{\textbf{ Illustration of the proposed Dual Attention Fusion.} }
    \label{attn}
\end{figure}

\subsection{Dual Attention Fusion (DAF)}

Figure~\ref{attn} present the structural details of the Dual Attention Fusion (DAF) unit within the DualSSM module. The DAF module is designed to adaptively fuse features extracted from both the global and local branches, thereby fully leveraging the strengths of each. This unit comprises two complementary attention branches: Global Stream Attention and Local Stream Attention, which operate on global and local features, respectively.

For the feature map obtained from global scanning, $Y_{\text{global}}\in\mathbb{R}^{H\times W\times D}$, we apply Global Stream Attention to recalibrate the response intensity of each channel. Specifically, we perform both average pooling and max pooling on $Y_{\text{global}}$ along the spatial dimensions. These two channel vectors are then fed into a two-layer MLP and add element by element to produce attention weights for each channel, $\mathbf{w}_c = [w_1,\dots,w_D] = \sigma(W_2(\delta(W_1(\mathbf{u}))))$, where $\delta(\cdot)$ denotes the ReLU activation function, $W_1\in \mathbb{R}^{D\times D/r}$ and $W_2\in \mathbb{R}^{D/r\times D}$ are the weight matrices of the two fully connected layers, $r$ is the channel reduction ratio (set to 4 by default), and $\sigma$ is the sigmoid activation function to ensure the output weights fall within the range of 0 to 1. The resulting channel weights $\mathbf{w}c$ are multiplied with the original feature map in a channel-wise manner, thus reweighting the channels of the global feature map. The output is the channel-enhanced global feature $Y_{\text{global}}' = \mathbf{w}c \odot Y_{\text{global}}$.

For the feature map obtained from local scanning, $Y_{\text{local}}\in\mathbb{R}^{H\times W\times D}$, we apply Local Stream Attention to highlight the importance of specific spatial locations. First, we perform both average pooling and max pooling along the channel dimension of $Y_{\text{local}}$, resulting in two single-channel feature maps of size $H\times W$, denoted as $M_{\text{avg}}$ and $M_{\text{max}}$. These two maps are then concatenated along the channel dimension to form a two-channel feature map $M_{\text{cat}}\in\mathbb{R}^{2\times H\times W}$. A $7\times7$ convolutional layer followed by a sigmoid activation is applied to $M_{\text{cat}}$, producing the spatial attention map $M_s=\sigma\big(f_{7\times7}(M_{\text{cat}})\big)\in\mathbb{R}^{1\times H\times W}$. Each element in $M_s$ takes a value between 0 and 1, indicating the importance of the corresponding spatial position. Finally, $M_s$ is broadcast across all $D$ channels and multiplied element-wise with the local features to obtain the spatially enhanced local feature map: $Y_{\text{local}}' = M_s \odot Y_{\text{local}}$.

After computing both channel attention and spatial attention, we sum the enhanced global feature $Y_{\text{global}}'$ and the enhanced local feature $Y_{\text{local}}'$ element-wise to obtain the fused feature map:

\begin{equation}
	Y_{\text{fused}} = Y'_{\text{global}} + Y'_{\text{local}} \label{yfuse}
\end{equation}

Subsequently, $Y_{\text{fused}}$ is normalized using LayerNorm to produce the final output of the DualSSM module. It is important to emphasize that, unlike simple fusion methods that combine global and local features using fixed weighted summation, the DAF module achieves a more fine-grained and dynamic fusion by applying attention mechanisms along both the channel and spatial dimensions to global and local information, respectively. This fusion strategy ensures that both global semantics and local details are preserved and enhanced, thereby improving the model's ability to represent complex visual patterns.

\subsection{Auxiliary Loss}

In the DualSSM module, both the global and local branches are each equipped with an auxiliary classification loss (indicated by gear icons in the figure), implemented via auxiliary classification heads. Each auxiliary head consists of a global average pooling layer followed by a fully connected layer, mapping the corresponding branch’s feature map to an output of the same dimension as the main task. Taking the $l$-th DualSSM module as an example, let the global branch output be reduced via GAP to $\mathbf{v}_g^{(l)}$, and the local branch output via GAP to $\mathbf{v}_l^{(l)}$. The auxiliary heads produce predictions $\hat{\mathbf{y}}_{g}^{(l)} = \text{FC}(\mathbf{v}_g^{(l)})$ and $\hat{\mathbf{y}}_{l}^{(l)} = \text{FC}(\mathbf{v}_l^{(l)})$, respectively. During training, the cross-entropy loss $L_{\text{aux}}$ is computed between each auxiliary prediction and the ground-truth label. These auxiliary losses are jointly optimized with the primary loss, and the total loss is defined as:

\begin{equation}
	L_{\text{total}} = L_{\text{main}} + \lambda \sum_{l} \left( L_{\text{aux},g}^{(l)} + L_{\text{aux},l}^{(l)} \right) \label{loss}
\end{equation}

where $L_{\text{main}}$ denotes the loss of the main classification head, while $L_{\text{aux},g}^{(l)}$ and $L_{\text{aux},l}^{(l)}$ represent the auxiliary losses from the global and local heads of the $l$-th module, respectively. The parameter $\lambda$ is a balancing coefficient, which is set to $\lambda=0.3$ in our experiments. It is worth noting that the auxiliary classification heads are used only during the training phase; during inference, these auxiliary branches are removed, introducing no additional computational overhead at test time.

\subsection{Architecture Variants}

We designed several versions of RoadMamba, namely RoadMamba-T, RoadMamba-S, and RoadMamba-B. The parameter counts of these models are comparable to mainstream Mamba-based architectures like VMamba-T, VMamba-S, and VMamba-B. Details of the model configurations can be found in the appendix. The main distinctions among these variants lie in the channel dimensions $C$ and the quantity of blocks within each stage.

\section{Experiments}

This section provides an overview of our experimental evaluation. The proposed model was extensively tested on the RSCD dataset and compared in detail with other state-of-the-art models. We also conducted ablation studies to thoroughly assess the contributions of each component in our approach. Furthermore, we discuss the rationale behind adopting a pure Mamba architecture instead of employing window-based attention mechanisms, such as those used in the Swin Transformer, for local feature extraction.

\subsection{Datasets}

The RSCD dataset addresses this shortcoming by providing a large-scale, richly annotated road image dataset, initially comprising 370,000 images and later expanded to one million, covering diverse road materials, usage years, traffic volumes, and a variety of seasonal, weather, and lighting scenarios. RSCD specifically labels road friction (dry, wet, water, fresh snow, melted snow, ice), material type (asphalt, concrete, mud, gravel), and surface unevenness (smooth, slightly uneven, severely uneven), resulting in 27 combined classes. This comprehensive annotation enables more robust research and development in road surface condition recognition for intelligent driving applications.

\subsection{Evaluation metrics}

To evaluate classification performance, we adopt several widely-used metrics, including Top-1 Accuracy, Mean Precision, Mean Recall, and Mean F1 Score. Top-1 Accuracy indicates the percentage of instances where the model's highest-confidence prediction matches the ground truth, serving as a standard indicator for single-label classification. Precision quantifies the ratio of correctly predicted positive samples to all samples predicted as positive, while Recall measures the proportion of true positives identified among all actual positive cases. These metrics together offer a comprehensive assessment of the model’s classification capability. Given the class imbalance present in the dataset, the F1 Score is also reported as it provides a harmonic mean of Precision and Recall, ensuring a more balanced evaluation.

\subsection{Implementation Details}

All experiments were performed on a single RTX 4090 GPU with a batch size of 32. Model optimization utilized AdamW, following a linear scaling rule to adapt the learning rate to different batch sizes, thus promoting training stability. Additional settings include a weight decay of $0.05$, an $\varepsilon$ value of $1\times10^{-8}$ for numerical stability, and momentum coefficients $\beta_1=0.9$ and $\beta_2=0.999$. The learning rate was scheduled using cosine annealing with an initial linear warmup phase.

\subsection{Comparative Experiments}

\begin{table}[h]
	\centering
	\caption{Comparison Experiments between RoadMamba and other SOTA models.} 
	\small
	% 增加行距
	\renewcommand{\arraystretch}{1.2}
	\setlength{\tabcolsep}{1pt}
    % \begin{adjustbox}{width=.49\textwidth} 
	\begin{tabular}{lcccccc}
		\toprule
		Model & Top-1 Acc & Mean-P & Mean-R & Mean-F1 & Params & GFLOPs\\
		\midrule
		  ConvNeXt-T  & 82.27 & 73.35 & 68.35 & 70.27 & 29M & 4.5 \\
            ConvNeXt-S  & 84.08 & 76.35 & 70.24 & 72.59 & 50M & 8.7\\
            ConvNeXt-B  & 84.08 & 75.55 & 70.50 & 72.48 & 89M & 15.4\\
            ConvNeXt-L  & 84.88 & 76.65 & 72.36 & 74.20 & 198M & 34.4\\
            \midrule
		  Swin-T  & 85.39 & 77.52 & 72.52 & 74.55 & 28M & 4.5\\
            Swin-S  & 85.16 & 77.21 & 72.44 & 74.44 & 50M & 8.7\\
            Swin-B  & 85.68 & 77.69 & 73.24 & 75.08 & 88M & 15.4\\
            Swin-L & 87.38 & 80.47 & 76.03 & 77.91 & 197M & 34.4\\
            \midrule
            ViT-B  & 86.83 & 78.31 & 74.99 & 76.38 & 86M & 18.5\\
            ViT-L  & 88.47 & 79.99 & 77.51 & 78.58 & 307M & 63.3\\
            \midrule
            VMamba-T  & 90.79 & 83.12 & 79.45 & 81.04 & 30M & 4.9\\
            VMamba-S  & 90.66 & 83.07 & 79.28 & 80.90 & 50M & 8.7\\
            VMamba-B  & 91.11 & 83.52 & 80.11 & 81.60 & 89M & 15.4\\
            \midrule
            LocalMamba-T  & 91.68 & 84.41 & 81.14 & 82.58 & 26M & 9.7\\
            LocalMamba-S  & 92.05 & 84.62 & 81.23 & 82.63 & 50M & 11.0\\
            \midrule
            MambaVision-T  & 91.52 & 84.10 & 80.77 & 82.23 & 32M & 4.4\\
            MambaVision-S  & 91.55 & 84.18 & 81.48 & 82.67 & 50M & 7.5\\
            MambaVision-B  & 92.07 & 84.99 & 82.23 & 83.49 & 98M & 15.0\\
            MambaVision-L  & 92.40 & 85.38 & 82.51 & 83.82 & 227M & 34.9\\
            % MaxViT-T     & 92.08 & 85.10 & 82.61 & 83.76 & 29M \\
            % FasterViT 0  & 91.45 & 84.42 & 81.25 & 82.66 & 31M \\
            \midrule
            \rowcolor{gray!15}
            RoadMamba-T  & 92.58 & 85.84 & 83.05 & 84.31 & 28M & 5.1\\
            \rowcolor{gray!15}
            RoadMamba-S  & 92.63 & 85.84 & 83.42 & 84.53 & 43M & 8.7\\
            \rowcolor{gray!15}
            RoadMamba-B  & 92.81 & 85.92 & 83.73 & 84.79 & 86M & 15.8\\

            % MaxViT-S     & 92.19 & 85.09 & 83.10 & 84.03 & 68M \\
            % FasterViT 1  & 92.02 & 85.00 & 82.26 & 83.51 & 54M \\
            % MaxViT-B     & 92.69 & 85.90 & 83.56 & 84.64 & 118M \\
            % FasterViT 2  & 92.28 & 85.39 & 82.57 & 83.85 & 76M \\
            % MaxViT-L     & 92.76 & 85.89 & 83.79 & 84.76 & 211M \\
            % FasterViT 3  & 92.13 & 85.11 & 82.56 & 83.70 & 160M \\
		\bottomrule
	\end{tabular}
    % \end{adjustbox}
    \label{compare}
\end{table}

Our comparative analysis evaluates the performance of the proposed method against several SOTA techniques. These methods include ConvNeXt \cite{liu2022convnet}, Swin Transformer \cite{liu2021swin}, Vision Transformer \cite{dosovitskiy2020image}, VMamba \cite{liu2024vmamba}, MambaVision \cite{hatamizadeh2025mambavision} and LocalMamba \cite{huang2024localmamba}.
In Table\ref{compare}, we present the RSCD classification re
sults. Each model is trained on 40 epochs. Specifically, We compare CNN-based models, Transformer-based models, CNN-Transformer hybrid architecture-based models, and Mamba-based models. Taking the model with the base size as an example, in terms of the Top-1 Accuracy, RoadMamba-B(92.81\%) achieves higher accuracy compared to ConvNeXt-B(84.08\%), Swin-B(85.68\%), ViT-B(86.83\%), VMamba-B(91.11\%) and MambaVision-B(92.07\%). Under the remaining three evaluation metrics, our approach also surpasses the others. This indicates that the RoadMamba proposed by us is an effective and promising paradigm.

\subsection{Why Choose Mamba?}

\begin{table}[t]
	\centering
	\caption{Performance comparison between Swin Transformer and WindowMamba.} 
	\small
	% 增加行距
	\renewcommand{\arraystretch}{1.2}
	\setlength{\tabcolsep}{2pt}
	\begin{tabular}{ccccc}
		\toprule
		Model & Top-1 Acc & Mean-P & Mean-R & Mean-F1 \\
		\midrule
		Swin-T    & 85.39 & 77.52 & 72.52 & 74.55 \\
		WindowMamba-T    & 89.22 & 80.08 & 75.46 & 77.31 \\
		\bottomrule
	\end{tabular} \label{why}
\end{table}

In designing our global-local dual-stream Mamba framework, a natural consideration was whether to incorporate window-based attention mechanisms, which have achieved remarkable success in architectures such as Swin Transformer. To rigorously justify our choice of adopting a pure Mamba architecture instead of integrating window-based attention, we conducted controlled comparative experiments. Specifically, we evaluated the performance of a window-based variant of Mamba (WindowMamba), using an identical window size as in Swin Transformer, and compared its performance directly against Swin Transformer.

In Table\ref{why} experimental results clearly demonstrated the superiority of WindowMamba over Swin Transformer, achieving notably higher accuracy metrics on the same dataset and task settings. 

Consequently, our findings reinforce the choice of a purely Mamba-based approach. The improved accuracy coupled with reduced computational complexity underscores Mamba’s advantage as a robust alternative to attention-based mechanisms, thus aligning with our goal of developing an efficient yet powerful model for fine-grained classification tasks.

\subsection{Ablation Study}

In this section, we present a set of ablation experiments to verify the effectiveness of the dual scanning strategy, DAF and auxiliary loss. Each model is trained on 20 epochs. More ablation experiments are presented in the appendix.

\begin{table}
	\centering
	\caption{Ablation Experiment,  A represents the dual scanning strategy, B represents DAF, C represents Auxiliary Loss.} 
	\small
	% 增加行距
	\renewcommand{\arraystretch}{1.2}
	\setlength{\tabcolsep}{2pt}
	\begin{tabular}{ccccc}
		\toprule
		Model & Top-1 Acc & Mean-P & Mean-R & Mean-F1 \\
		\midrule
		Without A\&B\&C & 90.46 & 82.59 & 78.65 & 80.35 \\
            Without B\&C    & 90.62 & 82.71 & 79.30 & 80.81 \\
            Without C    & 91.35 & 83.86 & 80.40 & 81.91 \\
            \rowcolor{gray!15}
            RoadMamba    & 91.52 & 84.35 & 80.95 & 82.44 \\
		\bottomrule
	\end{tabular} \label{ablation}
\end{table}

As shown in Table\ref{ablation}, to thoroughly investigate the contribution of each proposed component within our architecture, we conducted comprehensive ablation experiments. Specifically, we evaluated the individual impacts of the dual scanning strategy, DAF and Auxiliary Loss on the model's overall performance.

We first evaluated the baseline model that excluded all three proposed modules, corresponding to a Mamba architecture with only global scanning. Introducing the dual scanning mechanism resulted in a 0.16\% improvement in Top-1 accuracy. Further incorporating the Dual Attention Fusion led to an additional 0.73\% increase. Finally, adding the auxiliary loss yielded a further 0.17\% gain in Top-1 accuracy.

Overall, the ablation studies demonstrate the individual and collective importance of each component in our proposed framework, confirming their essential roles in achieving superior fine-grained classification performance.

\section{Conclusion}

In summary, this work pioneers the application of visual Mamba-based architectures to the field of road surface classification. We introduced RoadMamba, a novel framework that seamlessly integrates both local and global feature perception through the Dual State Space Model (DualSSM) and an effective Dual Attention Fusion (DAF) mechanism. By explicitly incorporating an auxiliary loss, our approach ensures balanced learning in the dual branches, mitigating the risk of over-reliance on global semantic cues and promoting the retention of critical local texture information. Extensive experiments conducted on RSCD dataset with one million road surface images demonstrate that RoadMamba sets a new benchmark for performance in this domain, confirming the potential of Mamba architectures for fine-grained scene understanding tasks.

% \section{Acknowledgments}

\bibliography{aaai2026}

\begin{table}[b]
	\centering
	\caption{Supplemental Ablation Experiment A,  GlobalMamba represents the strategy only uses Global Scan, WindowMamba represents the strategy only uses Local Scan.} 
	\small
	% 增加行距
	\renewcommand{\arraystretch}{1.2}
	\setlength{\tabcolsep}{2pt}
	\begin{tabular}{ccccc}
		\toprule
		Model & Top-1 Acc & Mean-P & Mean-R & Mean-F1 \\
		\midrule
		GlobalMamba    & 90.46 & 82.59 & 78.65 & 80.35 \\
            WindowMamba    & 89.22 & 80.08 & 75.46 & 77.31 \\
            \rowcolor{gray!15}
            RoadMamba*     & 90.62 & 82.71 & 79.30 & 80.81 \\
		\bottomrule
	\end{tabular} \label{ablation1}
\end{table}

\begin{table}
	\centering
	\caption{Supplemental Ablation Experiment B,  GLTC represents both global and local features go through the Channel Aggregator and Token Aggregator, GTLC represents global feature goes through the Token Aggregator and local feature goes through the Channel Aggregator, GCLT represents global feature goes through the Channel Aggregator and local feature goes through the Token Aggregator.} 
	\small
	% 增加行距
	\renewcommand{\arraystretch}{1.2}
	\setlength{\tabcolsep}{2pt}
	\begin{tabular}{ccccc}
		\toprule
		Model & Top-1 Acc & Mean-P & Mean-R & Mean-F1 \\
		\midrule
		GLTC & 91.29 & 83.85 & 80.35 & 81.87 \\
            GTLC & 91.12 & 83.49 & 80.08 & 81.59 \\
            \rowcolor{gray!15}
            GCLT & 91.35 & 83.86 & 80.40 & 81.91 \\
		\bottomrule
	\end{tabular} \label{ablation2}
\end{table}

\begin{table*}
	\centering
	\caption{Architectural overview of the RoadMamba series.}
	\label{tab:arch_RoadMamba}
	\small
	\renewcommand{\arraystretch}{1.2}
	\setlength{\tabcolsep}{2pt}
	\begin{tabular}{c c c c c}
		\toprule
		layer name & output size & RoadMamba-T &RoadMamba-S & RoadMamba-B \\
		\midrule
		stem & 112$\times$112 &
		conv 4$\times$4, 96, stride 4 &
		conv 4$\times$4, 96, stride 4 &
		conv 4$\times$4, 128, stride 4 \\
		\midrule
		stage 1 & 56$\times$56 &
		\begin{tabular}{@{}c@{}}
			DualSSM Block\\[-4pt]
			{\scriptsize$\left[
				\begin{array}{@{}l@{}}
					\textbf{Linear}~96{\rightarrow}384\\[-2pt]
					\textbf{DWConv}~3{\times}3,192\\[-2pt]
					\textbf{SS2D(Global Scan)},~192\\[-2pt]
					\textbf{SS2D(Local Scan)},~192\\[-2pt]
					\textbf{Multiplicative}\\[-2pt]
					\textbf{Linear}~192{\rightarrow}96
				\end{array}\right]$}$\times$1
		\end{tabular} &
		\begin{tabular}{@{}c@{}}
			DualSSM Block\\[-4pt]
			{\scriptsize$\left[
				\begin{array}{@{}l@{}}
					\textbf{Linear}~96{\rightarrow}384\\[-2pt]
					\textbf{DWConv}~3{\times}3,192\\[-2pt]
					\textbf{SS2D(Global Scan)},~192\\[-2pt]
					\textbf{SS2D(Local Scan)},~192\\[-2pt]
					\textbf{Multiplicative}\\[-2pt]
					\textbf{Linear}~192{\rightarrow}96
				\end{array}\right]$}$\times$3
		\end{tabular} &
		\begin{tabular}{@{}c@{}}
			DualSSM Block\\[-4pt]
			{\scriptsize$\left[
				\begin{array}{@{}l@{}}
					\textbf{Linear}~128{\rightarrow}512\\[-2pt]
					\textbf{DWConv}~3{\times}3,256\\[-2pt]
					\textbf{SS2D(Global Scan)},~256\\[-2pt]
					\textbf{SS2D(Local Scan)},~256\\[-2pt]
					\textbf{Multiplicative}\\[-2pt]
					\textbf{Linear}~256{\rightarrow}128
				\end{array}\right]$}$\times$3
		\end{tabular} \\
		& & patch merging$\rightarrow$192 & patch merging$\rightarrow$192 & patch merging$\rightarrow$256 \\
		\midrule
		stage 2 & 28$\times$28 &
		\begin{tabular}{@{}c@{}}
			DualSSM Block\\[-4pt]
			{\scriptsize$\left[
				\begin{array}{@{}l@{}}
					\textbf{Linear}~192{\rightarrow}768\\[-2pt]
					\textbf{DWConv}~3{\times}3,384\\[-2pt]
					\textbf{SS2D(Global Scan)},~384\\[-2pt]
					\textbf{SS2D(Local Scan)},~384\\[-2pt]
					\textbf{Multiplicative}\\[-2pt]
					\textbf{Linear}~384{\rightarrow}192
				\end{array}\right]$}$\times$3
		\end{tabular} &
		\begin{tabular}{@{}c@{}}
			DualSSM Block\\[-4pt]
			{\scriptsize$\left[
				\begin{array}{@{}l@{}}
					\textbf{Linear}~192{\rightarrow}768\\[-2pt]
					\textbf{DWConv}~3{\times}3,384\\[-2pt]
					\textbf{SS2D(Global Scan)},~384\\[-2pt]
					\textbf{SS2D(Local Scan)},~384\\[-2pt]
					\textbf{Multiplicative}\\[-2pt]
					\textbf{Linear}~384{\rightarrow}192
				\end{array}\right]$}$\times$3
		\end{tabular} &
		\begin{tabular}{@{}c@{}}
			DualSSM Block\\[-4pt]
			{\scriptsize$\left[
				\begin{array}{@{}l@{}}
					\textbf{Linear}~256{\rightarrow}1024\\[-2pt]
					\textbf{DWConv}~3{\times}3,512\\[-2pt]
					\textbf{SS2D(Global Scan)},~512\\[-2pt]
					\textbf{SS2D(Local Scan)},~512\\[-2pt]
					\textbf{Multiplicative}\\[-2pt]
					\textbf{Linear}~512{\rightarrow}256
				\end{array}\right]$}$\times$3
		\end{tabular} \\
		& & patch merging$\rightarrow$384 & patch merging$\rightarrow$384 & patch merging$\rightarrow$512 \\
		\midrule
		stage 3 & 14$\times$14 &
		\begin{tabular}{@{}c@{}}
			DualSSM Block\\[-4pt]
			{\scriptsize$\left[
				\begin{array}{@{}l@{}}
					\textbf{Linear}~384{\rightarrow}1536\\[-2pt]
					\textbf{DWConv}~3{\times}3,768\\[-2pt]
					\textbf{SS2D(Global Scan)},~768\\[-2pt]
					\textbf{SS2D(Local Scan)},~768\\[-2pt]
					\textbf{Multiplicative}\\[-2pt]
					\textbf{Linear}~768{\rightarrow}384
				\end{array}\right]$}$\times$6
		\end{tabular} &
		\begin{tabular}{@{}c@{}}
			DualSSM Block\\[-4pt]
			{\scriptsize$\left[
				\begin{array}{@{}l@{}}
					\textbf{Linear}~384{\rightarrow}1536\\[-2pt]
					\textbf{DWConv}~3{\times}3,768\\[-2pt]
					\textbf{SS2D(Global Scan)},~768\\[-2pt]
					\textbf{SS2D(Local Scan)},~768\\[-2pt]
					\textbf{Multiplicative}\\[-2pt]
					\textbf{Linear}~768{\rightarrow}384
				\end{array}\right]$}$\times$10
		\end{tabular} &
		\begin{tabular}{@{}c@{}}
			DualSSM Block\\[-4pt]
			{\scriptsize$\left[
				\begin{array}{@{}l@{}}
					\textbf{Linear}~512{\rightarrow}2048\\[-2pt]
					\textbf{DWConv}~3{\times}3,1024\\[-2pt]
					\textbf{SS2D(Global Scan)},~1024\\[-2pt]
					\textbf{SS2D(Local Scan)},~1024\\[-2pt]
					\textbf{Multiplicative}\\[-2pt]
					\textbf{Linear}~1024{\rightarrow}512
				\end{array}\right]$}$\times$15
		\end{tabular} \\
		& & patch merging$\rightarrow$768 & patch merging$\rightarrow$768 & patch merging$\rightarrow$1024 \\
		\midrule
		stage 4 & 7$\times$7 &
		\begin{tabular}{@{}c@{}}
			DualSSM Block\\[-4pt]
			{\scriptsize$\left[
				\begin{array}{@{}l@{}}
					\textbf{Linear}~768{\rightarrow}3072\\[-2pt]
					\textbf{DWConv}~3{\times}3,1536\\[-2pt]
					\textbf{SS2D(Global Scan)},~1536\\[-2pt]
					\textbf{SS2D(Local Scan)},~1536\\[-2pt]
					\textbf{Multiplicative}\\[-2pt]
					\textbf{Linear}~1536{\rightarrow}768
				\end{array}\right]$}$\times$3
		\end{tabular} &
		\begin{tabular}{@{}c@{}}
			DualSSM Block\\[-4pt]
			{\scriptsize$\left[
				\begin{array}{@{}l@{}}
					\textbf{Linear}~768{\rightarrow}3072\\[-2pt]
					\textbf{DWConv}~3{\times}3,1536\\[-2pt]
					\textbf{SS2D(Global Scan)},~1536\\[-2pt]
					\textbf{SS2D(Local Scan)},~1536\\[-2pt]
					\textbf{Multiplicative}\\[-2pt]
					\textbf{Linear}~1536{\rightarrow}768
				\end{array}\right]$}$\times$3
		\end{tabular} &
		\begin{tabular}{@{}c@{}}
			DualSSM Block\\[-4pt]
			{\scriptsize$\left[
				\begin{array}{@{}l@{}}
					\textbf{Linear}~1024{\rightarrow}4096\\[-2pt]
					\textbf{DWConv}~3{\times}3,2048\\[-2pt]
					\textbf{SS2D(Global Scan)},~2048\\[-2pt]
					\textbf{SS2D(Local Scan)},~2048\\[-2pt]
					\textbf{Multiplicative}\\[-2pt]
					\textbf{Linear}~2048{\rightarrow}1024
				\end{array}\right]$}$\times$3
		\end{tabular} \\
		\midrule
		& & \multicolumn{3}{c}{average pool, 27-d fc, softmax}\\
		\midrule
		Param. (M) & & 28 & 46 & 86\\
		FLOPs & & $5.1\times10^9$ & $8.7 \times10^9$ & $15.8\times10^9$ \\
		\bottomrule
	\end{tabular}
\end{table*}

\section{Appendix}

\subsection{Supplemental Ablation Experiments}

\subsubsection{Ablation Study on Scanning Strategies}
Table\ref{ablation1} presents the results of the supplemental ablation experiment comparing the performance of models employing only global scanning (GlobalMamba), only local scanning (WindowMamba), and the proposed dual-stream strategy (RoadMamba*). * represents that in this set of ablation experiments, Roadmamba only has a different scanning method from the previous two, and there is no increase in DAF module and auxiliary loss. As shown, RoadMamba consistently achieves the highest scores across all evaluation metrics, including Top-1 Accuracy, Mean Precision, Mean Recall, and Mean F1 Score. In contrast, both GlobalMamba and WindowMamba yield inferior results, indicating that relying solely on either global or local scan strategies limits the model's capacity for comprehensive feature extraction. These findings clearly demonstrate the effectiveness of RoadMamba's integration of both global and local scanning mechanisms, which enables more robust and accurate road surface classification.

\subsubsection{Ablation Study on Aggregator Assignment for Global and Local Streams}
Table\ref{ablation2} reports the results of the supplemental ablation experiment investigating different pathways for aggregating global and local features. Specifically, the results compare three configurations: GLTC (both global and local features pass through the Channel Aggregator and Token Aggregator), GTLC (global features use the Token Aggregator while local features use the Channel Aggregator), and GCLT (global features use the Channel Aggregator while local features use the Token Aggregator, i.e., the RoadMamba design). Among the three, the GCLT configuration achieves the best performance across all evaluation metrics, including Top-1 Accuracy, Mean Precision, Mean Recall, and Mean F1 Score. This demonstrates that assigning the Channel Aggregator to the global stream and the Token Aggregator to the local stream is the most effective strategy, highlighting the critical role of this architectural design in enabling RoadMamba to achieve superior road surface classification results.


\end{document}